\documentclass[conference,letterpaper]{IEEEtran}
\IEEEoverridecommandlockouts
\usepackage{cite}
\usepackage{amsmath,amssymb,amsfonts}
\usepackage{algorithmic}
\usepackage{graphicx}
\usepackage{textcomp}
\usepackage{xcolor}
\usepackage{url}  
\usepackage{multirow}
\usepackage{multicol}
\usepackage{xspace}
\def\etc{etc.\@\xspace}
\usepackage{verbatim}
\usepackage{subcaption}

\DeclareMathOperator{\charCNN}{charCNN}
\def\sm{\mathbf{S_m}}
\def\saux{\mathbf{S_{aux}^{k}}} 
\def\LK{\mathbf{L^k}}
\def\hak{\mathbf{h_a^k}}
\def\hmf{\mathbf{h_f^m}}
\def\hm{\mathbf{h^m}}
\def\zf{\mathbf{z_f}}
\def\gk{\mathbf{g_k}}
\def \org {\textit{School of Computing, Engineering and Built Environment}\\
\textit{Ulster University}\\
Belfast, UK
}

\def\BibTeX{{\rm B\kern-.05em{\sc i\kern-.025em b}\kern-.08em
    T\kern-.1667em\lower.7ex\hbox{E}\kern-.125emX}}
\begin{document}

\title{Gated Task Interaction Framework for Multi-task Sequence Tagging
}

\author{
\IEEEauthorblockN{Isaac K. E. Ampomah}
\IEEEauthorblockA{\org \\
ampomah-i@ulster.ac.uk}
\and
\IEEEauthorblockN{Sally McClean}
\IEEEauthorblockA{\org \\
si.mcclean@ulster.ac.uk}
\and
\IEEEauthorblockN{Zhiwei Lin}
\IEEEauthorblockA{\org \\
z.lin@ulster.ac.uk}
\and
\IEEEauthorblockN{Glenn Hawe }
\IEEEauthorblockA{\org \\
gi.hawe@ulster.ac.uk}
}

\maketitle

\begin{abstract}
Recent studies have shown that neural models can achieve high performance on several sequence labelling/tagging problems without the explicit use of linguistic features such as part-of-speech (POS) tags. 
These models are trained only using the character-level and the word embedding vectors as inputs.
Others have shown that linguistic features can improve the performance of neural models on tasks such as chunking and named entity recognition (NER).
However, the change in performance depends on the degree of semantic relatedness between the linguistic features and the target task; in some instances, linguistic features can have a negative impact on performance.
This paper presents an approach to jointly learn these linguistic features along with the target sequence labelling tasks with a new multi-task learning (MTL) framework called  Gated Tasks Interaction (GTI) network for solving multiple sequence tagging tasks. 
The GTI network exploits the relations between the multiple tasks via neural gate modules. These gate modules control the flow of information between the different tasks. Experiments on benchmark datasets for chunking and NER show that our framework outperforms other competitive baselines trained with and without external training resources.
\end{abstract}


\section{Introduction}
Neural approaches to sequence tagging problems such as \textit{Named Entity Recognition} (NER), \textit{Chunking} and \textit{Part-of-Speech} (POS) tagging have recently achieved remarkable performance \cite{collobert2011natural}. The
strength of neural models over traditional machine learning approaches such as Conditional Random Fields (CRF) \cite{lafferty2001conditional} and Hidden Markov Models (HMMs) \cite{zhou2002named} lies in their ability to automatically extract features from the input data. 
Recent studies \cite{huang2015bidirectional,lample2016neural} successfully trained a bidirectional long short-term memory (Bi-LSTM) with a CRF classifier based on the input word-based features
(mainly vector representation of words obtained from pre-trained word embedding vectors such as Glove \cite{pennington2014glove}). 
Character-level representations are also exploited to extend the word-based features as they enhance the performance of models with morphological information \cite{ma2016end,chiu2016named,peters2017semi,liu2017empower}. 

Aside from the word and the character-level representation, recent works  \cite{luo2017attention,sienvcnik2015adapting,marcheggiani2017encoding} suggest that higher performance can be obtained by leveraging additional information from linguistic features such as chunking and POS  tags. 
The POS and chunking features along with chemical were used as additional features to train a neural model for chemical NER \cite{luo2017attention}. 
They argued that the choice of additional linguistic information affects the performance on the NER task. 
Other recent works \cite{sogaard2016deep,hashimoto2017joint,strubell2018linguistically} also suggest directly modeling these linguistic features along with the target task in a multi-task learning fashion can further enhance the performance. 
~Hierarchical neural architectures for MTL were proposed by \cite{sogaard2016deep,hashimoto2017joint}. 
The multiple tasks were arranged in terms of their level of complexity to reflect the linguistic hierarchies of the tasks under consideration.
Finally, combining neural attention mechanism across dependency parsing, POS and predicate detection tasks, \cite{strubell2018linguistically} achieved a state-of-the-art performance on Semantic Role Labeling (SRL) task.
Though how to efficiently apply MTL to the task of sequence labelling has been extensively studied \cite{alonso2017multitask,gurevychoptimal,bingel2017identifying}, the performance of MTL depends on the relatedness of the tasks under consideration. Further research is required to investigate the design constraints and challenges to exploiting the relatedness between the main and auxiliary tasks under MTL efficiently and effectively.

To improve the performance of the multi-task semantic sequence tagging, this paper proposes a novel neural architecture, called Gated Task Interaction (GTI) network.
The GTI network takes a sequence of words as input and learns to generate the target sequence labels (e.g. NER) along with the linguistic features (such as the chunking and the POS tags).
The proposed GTI neural framework aims to capture the mutual dependencies between multiple sequence labeling problems. 
That is, given a main(target) task and its associated auxiliary tasks, the GTI model seeks to efficiently exploit the semantic relations between the given tasks so as to improve the model's performance in terms of learning the multiple objectives. 
To achieve high performance on the tasks, the GTI framework employs a gating mechanism to control the flow of information between the main task and its associated auxiliary tasks under consideration. 
The gating mechanism also serves as a regularization technique for improving the model's performance on the associated tasks \cite{zhang2016gated,xue2018aspect}. In this work, the auxiliary tasks and the main sub-networks are trained jointly end-to-end on the same dataset, eliminating the need for external datasets.  

The GTI framework is evaluated with two sequence labelling tasks: NER tagging on the CoNLL-2003 shared task \cite{tjong2003introduction}, and chunking on CoNLL-2000 shared task \cite{tjong2000introduction}.
~The contributions of this work are:
\begin{itemize}
\item proposing a novel neural architecture for multi-task linguistic sequence tagging/labelling.
\item evaluating the GTI framework empirically on benchmark datasets for sequence tagging problems (NER and chunking).
\item determining the impact of the choice of auxiliary tasks and hyperparameters (mainly the LSTM hidden state size) on the performance of our proposed approach.
\end{itemize}
The remainder of the paper is organized as follows. Section \ref{sec:gti_framework}  introduces our MTL sequence labeling framework.
The experiments conducted are presented in Section \ref{sec:experiment}, and the  results are compared and discussed in Section \ref{sec:res_disc}. Section \ref{sec:related_works} briefly discusses the related works. The conclusion is presented in Section \ref{sec:conc}.

\section{GTI Framework} \label{sec:gti_framework}
In this section, we provide a brief introduction to the components of our model and present the GTI framework for the multi-task sequence labelling problem.
 \begin{table}[t!]
        \center
        \caption{Notation Table.}
\label{tab:notates}
\scalebox{0.9}
{
\begin{tabular}{|l|l|l|l|}
\hline
Notation                   & \multicolumn{3}{c|}{Description}                                                           \\ \hline
$X$                        & \multicolumn{3}{l|}{model's input}                                                         \\ \hline
$x_i$                      & \multicolumn{3}{l|}{\emph{i}-th word/token}                                                \\ \hline
$\mathbf{x}$               & \multicolumn{3}{l|}{embedding of the model's input sequence}                               \\ \hline
$\mathbf{x_i}$             & \multicolumn{3}{l|}{embedding of the token/word $x_i$}                                     \\ \hline
$\mathbf{w}{(\cdot)}$      & \multicolumn{3}{l|}{word embedding look up table}                                          \\ \hline
$\mathbf{F}{(\cdot)}$      & \multicolumn{3}{l|}{`token/word format'  look up table}                                    \\ \hline
$\charCNN(\cdot)$          & \multicolumn{3}{l|}{character encoding sub-network}                                        \\ \hline
$\mathbf{w}{(x_i)}$        & \multicolumn{3}{l|}{word embedding of token $x_i$}                                         \\ \hline
$\mathbf{F}{(x_i)}$        & \multicolumn{3}{l|}{`token/word format' embedding for the i-th token}                      \\ \hline
$\charCNN(x_i)$            & \multicolumn{3}{l|}{character-level embedding  of the token $x_i$}                         \\ \hline
$\sm$                      & \multicolumn{3}{l|}{task specific semantic representation for the main task}               \\ \hline
$\saux$                    & \multicolumn{3}{l|}{task specific semantic representation for auxiliary task \emph{k}}            \\ \hline
$Y^k$                      & \multicolumn{3}{l|}{output label sequence under auxiliary task \emph{k}}                          \\ \hline
$Y^m$                      & \multicolumn{3}{l|}{output label sequence under the main task}                             \\ \hline
$\LK$                      & \multicolumn{3}{l|}{label embedding vector generated from $Y^k$}                           \\ \hline
$\gk$                      & \multicolumn{3}{l|}{gated auxiliary feature vector  from the auxiliary task \emph{k}} \\ \hline
$\mathbf{\hat{g}_k}$       & \multicolumn{3}{l|}{gate control vector for auxiliary task \emph{k}}                       \\ \hline
$\zf$       & \multicolumn{3}{l|}{ auxiliary feature vector generated from all the auxiliary task}                       \\ \hline
AuxTask\textsuperscript{k} & \multicolumn{3}{l|}{sub-network for auxiliary task \emph{k}}                                      \\ \hline
MT                         & \multicolumn{3}{l|}{the main task's sub-network}                                         \\ \hline
\end{tabular}
}
\end{table}
\begin{figure*}[t!]
	\centering
\begin{subfigure}{0.9\linewidth} 
		\includegraphics[width=\linewidth,keepaspectratio]{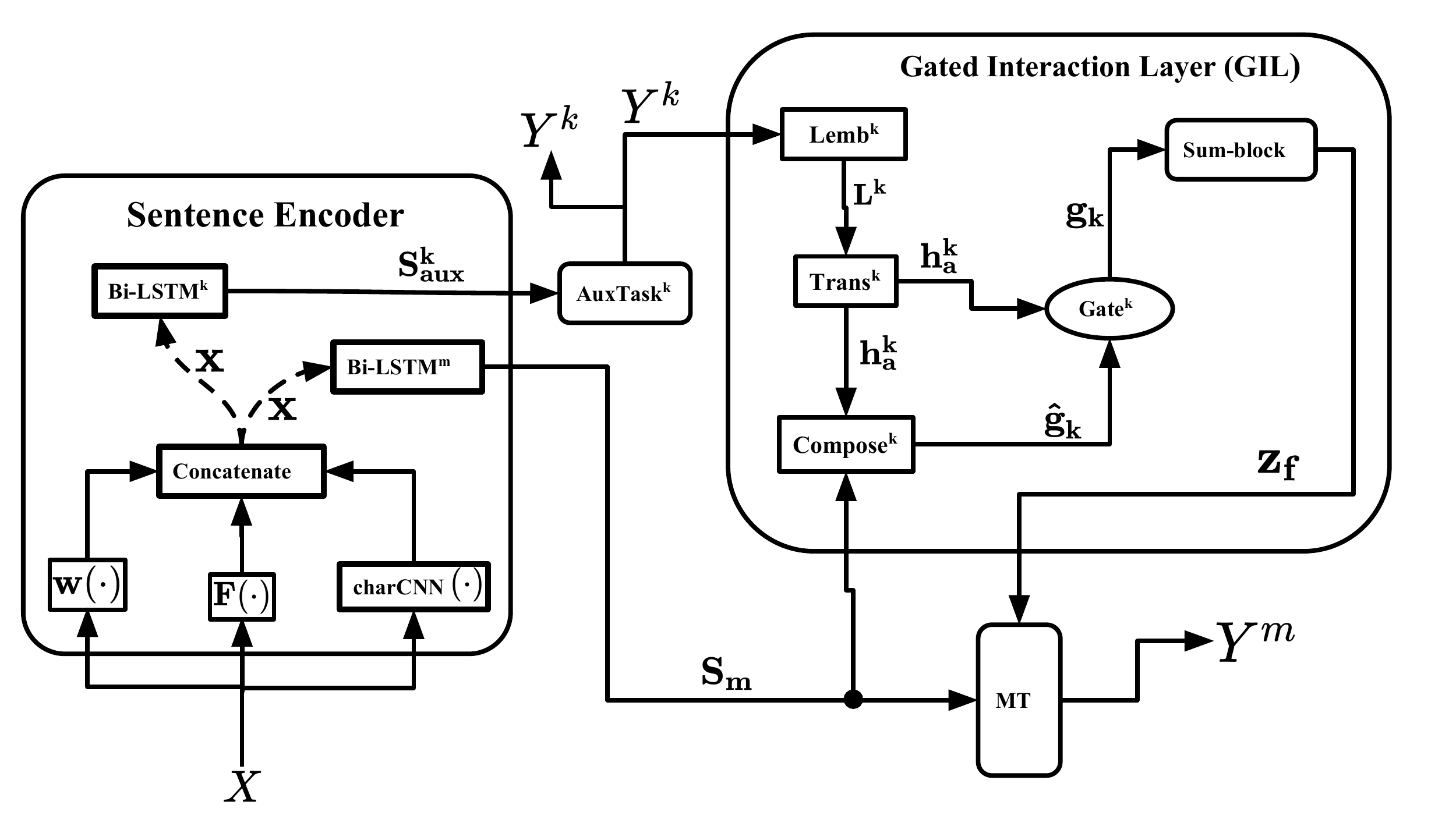}
		\caption{The basic form of the GTI architecture with a main-task $Y^m$ and one auxiliary task $Y^k$.} 
		\label{fig:gti_mtl_basic}
		\vspace*{2mm}
	\end{subfigure}
	\begin{subfigure}{0.9\linewidth} 
		\includegraphics[width=\linewidth,keepaspectratio]{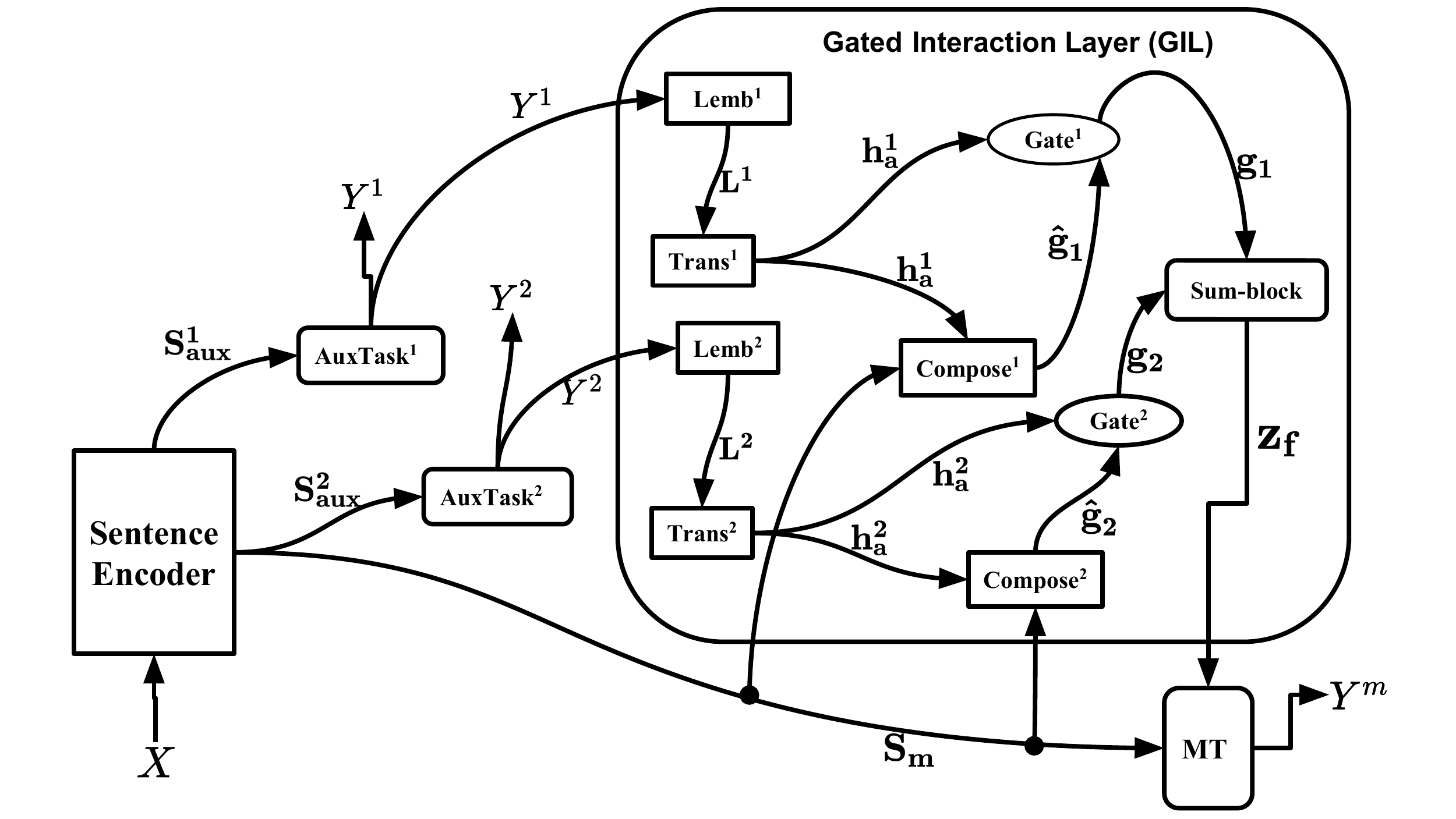}
		\caption{The GTI architecture with two auxiliary tasks sub-networks (\emph{AuxTask\textsuperscript{1}} and \emph{AuxTask\textsuperscript{2}} generating the output labels $Y^1$ and $Y^2$ respectively) and a single main/target task sub-network ( \emph{MT} generating the labels for the main Task $Y^m$).} 
		\label{fig:gti_mtl_v1}
		\vspace*{2mm}
	\end{subfigure}
	\caption{\small Gated Task Interaction (GTI) Architectures: \emph{AuxTask\textsuperscript{k}} ($\{k : 1\leq k\leq K\}$) and \emph{MT} are  the auxiliary and the main task sub-networks respectively, $\sm$ and $\saux$ are the task-specific representation vectors generated from the input token sequence $X$ for learning the main and auxiliary task \emph{k} respectively.  }
\label{fig:gti_mtl_models}
\end{figure*}
\subsection{Gated Task Interaction (GTI) Network}
Fig.~\ref{fig:gti_mtl_basic} illustrates the basic neural network architecture of the proposed GTI framework and our architecture with two auxiliary tasks ($Y^1$ and $Y^2$) is shown in Fig.~\ref{fig:gti_mtl_v1}.
Given an input sequence $X=[x_1,x_2, \cdots ,x_n]$ containing \emph{n} tokens, a main/target task $Y^m=\left[y^m_1, y^m_2,  \cdots, y^m_n \right]$ and a set of \emph{K} auxiliary tasks \{$Y^{k} : 1\leq k\leq K\}$, where $Y^k=[y^k_1, y^k_2,  \cdots, y^k_n ]$ is an auxiliary task, the GTI framework aims (1) to extract features from predicting $Y^k$, and (2) to use the extracted features for predicting $Y^m$. These notations are summarized in Table~\ref{tab:notates}.

Our GTI network is composed of the following  primary components:
\begin{enumerate}
    \item \textbf{Sentence Encoder:} handles the generation of the task specific semantic representations $\saux\in \mathbb{R}^{n\times d}$ and $\sm\in \mathbb{R}^{n\times d}$ for the auxiliary task \emph{k} and the main-task respectively from the input sequence $X$ (where \emph{n} is the number of tokens and \emph{d} is the dimension of the hidden representation).
    \item \textbf{AuxTask\textsuperscript{k} Network:}  handles the generation of the output tags $Y^k$ from the $\saux$ .
    \item \textbf{Gated Interaction Layer (GIL):} extracts an auxiliary feature vector $\zf$ using the  output tags 
    from the auxiliary tasks under consideration via the Lemb\textsuperscript{k}, the Trans\textsuperscript{k}, the Compose\textsuperscript{k}, the Gate\textsuperscript{k} and sum-blocks sub-modules.
    \item \textbf{Main-task (MT) Network:} generates main/target task output tags $Y^m$ from $\sm$ and $\zf$. 
\end{enumerate}
For sequence tagging problems such as NER, the tagging scheme imposes a constraint on the order of the output tags. For example, using the IOBES tagging scheme, output tag sequence starting with I- tag without a preceding B- tag or having I-MISC following a B-LOC is an illegal and meaningless sequence.
Softmax classifiers fail to capture the tag-order constraint imposed by the tagging scheme. 

Unlike the softmax classifier, CRF models the tagging decisions jointly hence capturing the dependencies between them \cite{lample2016neural,ma2016end}. ~Therefore we use CRF to predict the output tags for both the  AuxTask\textsuperscript{k} and Main-task (MT) networks. For each task under consideration, we compute the sentence CRF loss  \cite{lafferty2001conditional} using the forward-backward algorithm at training time. Following \cite{collobert2011natural}, we use the Viterbi algorithm to find the most likely tag sequence during testing.
\subsubsection*{\textbf{Sentence Encoder}}
The first stage of this layer generates the embedding $\mathbf{x_i} \in \mathbb{R}^{d}$ for each token $x_i$ as the concatenation of the word embedding $\mathbf{w}(x_i)$, the character-level embedding from $\charCNN(x_i)$ and the `token/word format' embedding $\mathbf{F}(x_i)$ as shown below:
\begin{equation}
    \mathbf{x_i} = \left[\mathbf{w}(x_i);\charCNN(x_i);\mathbf{F}(x_i)\right]
\end{equation}
$\mathbf{w}(\cdot)$ is a token look-up table which maps the token  $x_i$ to a high dimensional vector $\mathbf{w}(x_i)$. 
The word format embedding $\mathbf{F}(\cdot)$~(initialised as a 1-hot feature vector)  maps the `token-format' (i.e. whether it is numeric, punctuation, lower, upper-cased or alphanumeric) of  $x_i$ into a representation vector.
$\charCNN(\cdot)$ is a character  encoding  sub-network computing the embedding of an input word based on its characters.
As shown by \cite{liu2017empower,lample2016neural,ma2016end}, learning a word representation from its character captures important morphological information useful for solving several tagging tasks such as NER, POS \etc 
Within $\charCNN(\cdot)$, a look-up table first maps the sequence of characters of a given token $x_i$ into a vector sequence $\mathbf{C^i}=\left[\mathbf{c^i_1},\mathbf{c^i_2}, \cdots,\mathbf{c^i_l}\right]$, where $\mathbf{c^i_j}$ is the vector \textit{j}-th character in the \textit{i}-th token of length \emph{l}. The generated vector sequence is then passed to a character-CNN \cite{ma2016end,chiu2016named}, which computes the character-level embedding feature vector  of the input token $x_i$ after the convolution operations. 
The character look-up table is updated during training. 

The last stage of the sentence encoder is the generation of the contextual-semantic representation vectors ($\saux$ and $\sm$) from the input sequence embedding $\mathbf{x}$ as shown in Fig. \ref{fig:gti_mtl_basic}. $\saux$ and $\sm$ are the task-specific representation vectors computed to solve the auxiliary task \emph{k} and the main/target-task \emph{m} respectively. BiLSTM network is adopted to compute these vectors. $\saux$ and $\sm$ are obtained by concatenating the backward and the forward context representation vectors from their respective BiLSTM encoder. Dropouts of rate 0.25 is applied to the input of the all the BiLSTMs networks.
\subsubsection*{\textbf{AuxTask\textsuperscript{k} Network}}
This sub-network takes as input the auxiliary task-specific semantic representation $\saux$ and generate the label sequence $Y^k=[y^k_1,y^k_2, \cdots ,y^k_n ]$ for auxiliary task \emph{k}. 
The AuxTask\textsuperscript{k} network consists of a linear transformation layer to process the $\saux$ into a hidden representation $\mathbf{A^k} \in \mathbb{R}^{n \times t}$ (where \emph{n} is the number of tokens and \emph{t} is the number possible output tags under the auxiliary task \emph{k}) as shown in Eq. \eqref{eqn:transK}.
\begin{align}
    \mathbf{A^k} &= \mathbf{W^{k}_{A}}{\saux} + \mathbf{b_A^k} \label{eqn:transK}
\end{align}
where $\mathbf{W^{k}_{A}}$ and $\mathbf{b_A^k}$ are trainable parameters. The $\mathbf{A^k}$ is then passed to the CRF classifier which generates the $Y^k$ tags.
\subsubsection*{\textbf{Gated Interaction Layer (GIL)}}
This layer controls the flow of information between the AuxTask\textsuperscript{k} and the MT sub-networks. The GIL computes the auxiliary feature vector $\zf$ from the output of the AuxTask\textsuperscript{k} network and $\sm$ (computed by the sentence encoder).
As shown in Fig. \ref{fig:gti_mtl_basic}, it consists of the Lemb\textsuperscript{k}, the Trans\textsuperscript{k}, the Compose\textsuperscript{k}, the Gate\textsuperscript{k} and the sum-block layers.
Within the Lemb\textsuperscript{k}, the $Y^k$ (the one-best tagging sequence predicted by the AuxTask\textsuperscript{k} network) is  mapped into a label embedding vector $\LK$ using a randomly initialized lookup table.
The Trans\textsuperscript{k} consists of a single layer BiLSTM which processes the $\LK$ vector into $\hak$. 
Based on the $\hak$ vector and $\sm$,  the gate control vector $\{\mathbf{\hat{g}_k}: 1\leq k \leq K\}$ is generated by the Compose\textsuperscript{k} sub-layer as shown in Eq. \eqref{eqn_gates_k1}.
\begin{equation} \label{eqn_gates_k1}  
    \mathbf{\hat{g}_k} = \mathbf{W_k}{\hak} + \mathbf{U_k}{\sm}   
    \end{equation}
    where $\mathbf{U_k}$ and $\mathbf{W_k}$ are weight matrices for generating the $\mathbf{\hat{g}_k}$.
Given the $\mathbf{\hat{g}_k}$ and $\hak$ vectors, the Gate\textsuperscript{k}  generates the gated auxiliary feature vector $\gk$ for the auxiliary task \emph{k} as shown below:
\begin{equation} \label{eqn_gates_k2}
    \gk = \sigma\left(\mathbf{\hat{g}_k}\right) \odot \mathbf{{h^k_a}} 
\end{equation}
where $\odot$ represents an element-wise product operation and  $\sigma(\cdot)$ is the sigmoid activation function.
The final stage is the computation of $\zf$, which is a weighted summation of all the $\gk$ vectors.
This is done by the sum-block sub-module as shown in Eq. \eqref{eqn_aux}.
\begin{align}
    \zf &=  \mathbf{W_{f}}\sum_{k=1}^{K}\left({\gk} \right)   \label{eqn_aux}
\end{align}
Here $\mathbf{W_{f}}$ is a weight matrix for the generation of the $\zf$ and $K$ is the total number of auxiliary tasks under consideration.
\subsubsection*{\textbf{Main-task (MT) Network}}
This is the sub-network for generating the sequence labels $Y^m$ under the main task. It takes as input  the $\zf$ and the $\sm$ representation vectors from the GIL and the sentence encoder respectively. These input vectors are then combined to compute the representation vector $\hmf$ as shown in Eq. \eqref{eqn:hm}.
\begin{align}
    \hmf &= \tanh\left(\mathbf{W_m}{\sm} + \zf  \right) \label{eqn:hm}
\end{align}
where $\mathbf{W_m}$ is a weight matrix. 
A single BiLSTM layer processes the $\hmf$ into the hidden representation $\hm$ by concatenating the right context representation ($\mathbf{\overrightarrow{h}^m}$)  and left context representation ($\mathbf{\overleftarrow{h}^m}$) vectors, i.e. $\hm =[\mathbf{\overrightarrow{h}^m};\mathbf{\overleftarrow{h}^m}]$. A linear transformation layer then processes the $\hm$ into an intermediate representation vector $A_f^m \in \mathbb{R}^{n\times m}$ (where \emph{n} is the number of tokens and \emph{m} is the number possible output tags under the main task) as shown below:
\begin{align}
    \mathbf{A^m} &= \mathbf{W_A}{\hm} + \mathbf{b_A} \label{eqn:transM}
\end{align}
where $\mathbf{W_A}$ and $\mathbf{b_A}$ are trainable parameters.
A CRF classifier then generates the output tags $Y^m=[y_1^m,y_2^m, \cdots y_n^m]$ from the $\mathbf{A^m}$. 
\subsection{Joint Model Training}
A common approach to learning multiple objectives via MTL involves randomly selecting and alternating between the different tasks and their associated dataset \cite{hashimoto2017joint,alonso2017multitask,sogaard2016deep}. 
During the training instance \emph{t}, the loss function is computed only based on the current task's objective.
But in this work, the main-task and auxiliary tasks sub-networks are jointly trained on only the same  in-domain dataset. 

Therefore, given the CRF loss $L_m$ (computed by the main-task's network) and the losses from the AuxTask\textsuperscript{k} networks $L_{aux}$ (see Eq. \eqref{eqn_aux_loss}), we define the joint training loss function $J_{loss}$ as:
\begin{align}
     L_{aux}  &= \sum_{k=1}^{K} {L(\hat{y}^{(k)}, y^{(k)})} \label{eqn_aux_loss} \\
    J_{loss} &= L_m + L_{aux}
\end{align}
where $L(\hat{y}^{(k)}, y^{(k)})$ is the CRF loss computed from the auxiliary task \emph{k}. 

\section{Experiment}\label{sec:experiment}
We evaluate the performance of the proposed GTI framework on the CoNLL-2000 Chunking dataset \cite{tjong2000introduction} and CoNLL-2003 NER dataset \cite{tjong2003introduction}. We use the IOBES tagging scheme which as shown by \cite{gurevychoptimal} yields a better performance compared to the IOB and the BIO schemes. 
\begin{itemize}
\item \textbf{CoNLL-2000 Chunking} was generated from the Wall Street Journal (WSJ) corpus with sections 15-18 as the training, section 20 for testing. We randomly sampled  1000 sentences from the training set as the validation dataset. Each sentence is pre-labelled with the POS tags and the  syntactic chunk tags. Therefore the  experiment conducted on this dataset predicts the POS tags as the auxiliary task and chunk labels prediction as the main-task. We refer to the GTI network trained on this dataset as GTI-CHUNK-POS. 
\item \textbf{CoNLL-2003 NER} contains annotations for the following entity types: LOC, PER, ORG and MISC. Each sentence is also pre-labelled with the chunking and POS tags. The dataset comes with training, development and test set splits. Unlike the works by ~\cite{yang2017transfer,chiu2016named,peters2017semi}, we only trained our models on the training set and use the development set for parameter tuning and report the results in terms of the models' performance on the test set. We performed experiments with 3 different model configurations/instances on this dataset as shown in the Table \ref{tbl:ner_models}. These models were trained with the prediction of the appropriate POS and Chunk tags as the auxiliary tasks and the named entity labels as the main-task.
    \begin{table}[t!]
        \center
        \caption{Models for CoNLL-2003 NER task.}
\scalebox{0.9}
{
        \begin{tabular}{|l|l|l|l|}
        \hline
       Index & Model                      & Main/Target task & AuxTasks  \\\hline
        1 & GTI-NER-POS                   & NER              & POS             \\ \hline
        2 & GTI-NER-CHUNK                 & NER              & Chunking          \\ \hline
        3 & GTI-NER-CHUNK-POS             & NER              & Chunking and POS  \\ \hline

        \end{tabular}
        }
        
        \label{tbl:ner_models}
    \end{table}
\end{itemize}

\subsection{Model Setup and Hyperparameters}
\begin{itemize}
 \item \textbf{Initialization: }The word embedding vectors were initialized with pre-trained embedding vectors of dimension equal to 100 from Glove\footnote{\url{http://nlp.stanford.edu/projects/glove/}} \cite{pennington2014glove}. The out-of-vocabulary (OOV) words were initialized by uniform sampling from the distribution [-0.25,0.25].
During training, we do not update the embedding vectors. 
The character embedding was initialized by sampling from the uniform distribution [-0.5,0.5] with the dimension of the resulting vector equal to that of the pre-trained word embedding vectors. 
 \item \textbf{Training: } We set the dropout rates for all the LSTMs, character-CNN and the output of the Trans\textsuperscript{k} layers to 0.25.
 The models are trained with the batch-size set as 10.
We use the Nadam optimizer with the initial learning rate $\alpha_0$. 
During training, the learning rate is updated via  a shifted cosine annealing function proposed by \cite{loshchilov2016sgdr}. 
This approach anneals the learning rate from its initial value $\alpha_0$ to 0 during a single cosine cycle and  introduces extra parameters such as the total number of iterations (total number of epochs) \emph{T} and the number of annealing cycle \emph{M}. 
After several initial experiments to pick the best values of \emph{T}, \emph{M} and $\alpha_0$, we choose  $\alpha_0=0.001$, \emph{M}=9 and \emph{T}=270. 
We did not train the models for the entire \emph{T} but rather stopped the training 10 epochs after the first 2 cycles (i.e. each model is trained for a total of 70 epochs). 
This is because  we observed a marginal improvement in each model's performance after the first 2 cycles. 
The neural network architecture was  implemented using Keras (version 2.0.5 with tensorflow backend). The models were trained on GeForce GTX 1080 Ti 11 GB RAM GPU.
The code will be available at online\footnote{\url{https://github.com/kaeflint/GTI}}. 
\end{itemize}
There are several parameters whose value affects the performance of the proposed approach. These parameters include but not limited to the dropout rates, number of BiLSTM layers, initial learning rate, choice of RNN units (either LSTM or GRU).
In this work, we only explore the impact of the LSTM hidden state size. 
Therefore, keeping parameters such as dropout rate, dimension of the character and word-format embedding and learning rate the same across the different configurations of the GTI network, we train the models with different the LSTM hidden state size selected from the set $\{100,150,200\}$. 
We evaluate the performance on the NER and chunking prediction based on the official evaluation metric (micro-averaged F$_1$ score). 

\section{Results and Discussion}\label{sec:res_disc}
In this section, we discuss the performance of our GTI models. Under all the training scenarios, we observed that the GTI network's performance on the main-task depends on the auxiliary tasks handled by the  AuxTask\textsuperscript{k} sub-networks. The results (see Table~\ref{tab:ner_scores_state} and Table~\ref{tab:chunk_scores_state}) obtained suggest that aside the relatedness between the tasks, the dimension of the LSTM hidden state also affect the performance especially in the case of NER tagging. Increasing the hidden state size improves the performance at a small cost of computational overhead due to increasing size of the model. 
\begin{table}[t!]
\center
\caption{Effect of hidden state size of LSTM on the performance of the GTI models  on the CoNLL-2003 NER dataset.}
\label{tab:ner_scores_state}
\scalebox{0.9}
{
\begin{tabular}{|l|l|c|c|c|}
\hline
\multicolumn{1}{|c|}{\multirow{2}{*}{Model}} & \multicolumn{1}{c|}{\multirow{2}{*}{State size}} & \multicolumn{3}{c|}{F$_1$ score} \\ \cline{3-5} 
\multicolumn{1}{|c|}{}                       & \multicolumn{1}{c|}{}                            & min   & mean ($\pm$ \textbf{std}) & max   \\ \hline
\multirow{3}{*}{GTI-NER-POS}                 & 100                                              & 90.94 & 91.09$\pm$0.09  & 91.17 \\ \cline{2-5} 
                                             & 150                                              & 90.84 & 91.04$\pm$0.13  & 91.22 \\ \cline{2-5} 
                                             & 200                                              & 91.20 & 91.33$\pm$0.14   & 91.56 \\ \hline
\multirow{3}{*}{GTI-NER-CHUNK}               & 100                                              & 90.88 & 91.08$\pm$0.13   & 91.22 \\ \cline{2-5} 
                                             & 150                                              & 90.80 & 90.93$\pm$0.08   & 91.04 \\ \cline{2-5} 
                                             & 200                                              & 91.11 & 91.22$\pm$0.09   & 91.33 \\ \hline
\multirow{3}{*}{GTI-NER-CHUNK-POS}           & 100                                              & 90.83 & 90.86$\pm$0.02   & 90.89 \\ \cline{2-5} 
                                             & 150                                              & 91.05 & 91.18$\pm$0.10   & 91.30 \\ \cline{2-5} 
                                             & 200                                              & 91.27 & 91.38$\pm$0.07   & 91.49 \\ \hline
\end{tabular} }

\end{table}
\begin{table}[t!]
\center
\caption{\small Effect of hidden state size of LSTM on the performance of the GTI models  on the CoNLL-2000 Chunking dataset.}
\label{tab:chunk_scores_state}
\scalebox{0.9}
{
\begin{tabular}{|l|l|c|c|c|}
\hline
\multicolumn{1}{|c|}{\multirow{2}{*}{Model}} & \multicolumn{1}{c|}{\multirow{2}{*}{State size}} & \multicolumn{3}{c|}{F$_1$ score}                                                               \\ \cline{3-5} 
\multicolumn{1}{|c|}{}                       & \multicolumn{1}{c|}{}                            & min                        & mean ($\pm$ \textbf{std})                     & max                        \\ \hline
\multirow{3}{*}{GTI-CHUNK-POS}               & 100                                              & \multicolumn{1}{l|}{94.91} & \multicolumn{1}{l|}{94.96$\pm$0.04} & \multicolumn{1}{l|}{95.03} \\ \cline{2-5} 
                                             & 150                                              & \multicolumn{1}{l|}{95.01} & \multicolumn{1}{l|}{95.05$\pm$0.03} & \multicolumn{1}{l|}{95.11} \\ \cline{2-5} 
                                             & 200                                              & \multicolumn{1}{l|}{95.08} & \multicolumn{1}{l|}{95.15$\pm$0.05} & \multicolumn{1}{l|}{95.22} \\ \hline
\end{tabular} }

\end{table}
\begin{table}[t!]
\center
\caption{CoNLL-2003 test set F$_1$ score. $^\dagger$ indicates models trained with pre-trained word embedding.}
\label{tab:ner_results_comp}
\scalebox{0.9}
{
\begin{tabular}{|c|l|l|l|}
\hline
\multicolumn{1}{|l|}{External Resources}                                                                                    & Model                                                                           & \multicolumn{2}{l|}{\begin{tabular}[c]{@{}l@{}}F$_1$ $\pm $ \textbf{std}\end{tabular}} \\ \hline
\multicolumn{1}{|l|}{\multirow{2}{*}{gazetteers}}                                                                           & Collobert et al. 2011 \cite{collobert2011natural} $^\dagger$          & \multicolumn{2}{l|}{89.59}                                                             \\ \cline{2-4} 
\multicolumn{1}{|l|}{}                                                                                                      & Chiu et al. 2016 \cite{chiu2016named} $^\dagger$  & \multicolumn{2}{l|}{91.62$\pm$0.33}                                                    \\ \hline
\multicolumn{1}{|l|}{AIDA dataset}                                                                                          & Luo et al. 2015 \cite{luo2015joint}              & \multicolumn{2}{l|}{91.20}                                                             \\ \hline
\multicolumn{1}{|l|}{\begin{tabular}[c]{@{}l@{}}CoNLL2000/\\ PTB-POS\end{tabular}}                                          & Yang et al. 2017 \cite{yang2017transfer}$^\dagger$       & \multicolumn{2}{l|}{91.26}                                                             \\ \hline
\multicolumn{1}{|l|}{\begin{tabular}[c]{@{}l@{}}1B Word dataset \&\\ \texttt{4096-8192-1024}\end{tabular}} & \multirow{2}{*}{Peters et al. 2017 \cite{peters2017semi}$^\dagger$} & \multicolumn{2}{l|}{91.93$\pm$0.19}                                                    \\ \cline{1-1} \cline{3-4} 
\multicolumn{1}{|l|}{1B Word dataset}                                                                                       &                                                                                 & \multicolumn{2}{l|}{91.62$\pm$0.23}                                                    \\ \hline
\multirow{15}{*}{None}                                                                                                      
& Peters et al. 2017 \cite{peters2017semi}$^\dagger$                  & \multicolumn{2}{l|}{90.87$\pm$0.13}                                                    \\ \cline{2-4} 
& Collobert et al. 2011 \cite{collobert2011natural}$^\dagger$          & \multicolumn{2}{l|}{88.67}                                                             \\ \cline{2-4} 
& Chiu et al. 2016 \cite{chiu2016named} $^\dagger$                 & \multicolumn{2}{l|}{90.91$\pm$0.20}                                                    \\ \cline{2-4}                                                                                                                             & Luo et al. 2015 \cite{luo2015joint}                               & \multicolumn{2}{l|}{89.90}                                                             \\ \cline{2-4} 
                                                                                                                            
                                                                                                                            & Yang et al. 2017 \cite{yang2017transfer}$^\dagger$                  & \multicolumn{2}{l|}{91.20}                                                             \\ \cline{2-4} 
                                                                                                            & Rei 2017 \cite{rei2017semi} $^\dagger$                               & \multicolumn{2}{l|}{86.26}                                                             \\ \cline{2-4} 
                                                                                                                            & Lample et al. 2016 \cite{lample2016neural} $^\dagger$                 & \multicolumn{2}{l|}{90.94}                                                             \\ \cline{2-4} 
                                                                                                                            & Ma et al. 2016 \cite{ma2016end}$^\dagger$                          & \multicolumn{2}{l|}{91.21}                                                             \\ \cline{2-4} 
                                                                                                                  \cline{2-4}
                                                                                                                            & \multirow{2}{*}{GTI-NER-POS$^\dagger$}                                          & mean                                  & 91.33$\pm$0.14                                 \\ \cline{3-4} 
                                                                                                                            &                                                                                 & max                                   & 91.56                                          \\ \cline{2-4} 
                                                                                                                            & \multirow{2}{*}{GTI-NER-CHUNK$^\dagger$}                                        & mean                                  & 91.22$\pm$0.09                                 \\ \cline{3-4} 
                                                                                                                            &                                                                                 & max                                   & 91.33                                          \\ \cline{2-4} 
                                                                                                                            & \multirow{2}{*}{GTI-NER-CHUNK-POS$^\dagger$}                                    & mean                                  & 91.38$\pm$0.07                                 \\ \cline{3-4} 
                                                                                                                            &                                                                                 & max                                   & 91.49                                          \\ \hline
\end{tabular} 
}

\end{table}
\subsection{Performance on the NER Tagging}
The results of our GTI models and existing approaches for the NER task are summarised in Table \ref{tab:ner_results_comp}. 
To make a fair comparison, we report the F$_1$ scores of existing models with and without the use of external labelled corpus such as the gazetteers, AIDA and PTB-POS datasets. Our GTI models for the NER task are trained only on the CoNLL-2003 NER dataset. 
Among our models, the GTI-NER-CHUNK-POS achieved the overall best performance with a score of 91.49 (91.38 $\pm$ 0.07) approximately $0.44$ less than the current state-of-the-art model~\cite{peters2017semi}.
Their MTL model for NER tagging and neural language modelling tasks was trained along with a large amount of extra training corpus which resulted in a score of 91.93. 
Without any external resources, their model's F$_1$ score dropped to 90.87 representing a significant loss in performance. 
Aside from the model proposed by \cite{chiu2016named}, our GTI-NER-CHUNK-POS outperforms all other systems/models, including those trained using external corpus like gazetteers. 

Table \ref{tab:ner_scores_state} presents the impact of the hidden state size. One can clearly observe that in most cases increasing the LSTM's hidden state size produces a better performance on the NER tagging. Increasing the hidden state size enables the GTI models to effectively extract/encode the necessary information needed to enhance the performance on the NER task. In all the experiments on the CoNLL-2003 NER dataset, we observed that the choice of auxiliary task consistently affects the overall performance of the GTI models. Compared to chunking task, learning NER  along with the POS tagging as auxiliary task (GTI-NER-POS) produced the better model but at a cost of higher variance. On the other hand, among our models, jointly learning the NER tagging along with the linguistic features (chunking and POS tagging) produced the overall best performance  in terms of the mean F$_1$ score and it achieved lower variance. 
\subsection{Performance on Chunking}
\begin{table}[t!]
\center
    \caption{CoNLL-2000 Chunking test set F$_1$ score. }
    \label{tab:chunk_f1}
\scalebox{0.9}
{
\begin{tabular}{|c|l|l|l|}
\hline
\multicolumn{1}{|l|}{External Resources}       & Model                                   & \multicolumn{2}{l|}{F$_1$ $\pm $ \textbf{std}}               \\ \hline
\multicolumn{1}{|l|}{\multirow{2}{*}{PTB-POS}} & S{\o}gaard et al. 2016 \cite{sogaard2016deep} & \multicolumn{2}{l|}{95.56}                  \\ \cline{2-4} 
\multicolumn{1}{|l|}{}                         & Hashimoto et al. 2017                   & \multicolumn{2}{l|}{95.77}                  \\ \hline
\multicolumn{1}{|l|}{CoNLL2000/ PTB-POS}       & Yang et al. 2017 \cite{yang2017transfer} & \multicolumn{2}{l|}{95.41}                  \\ \hline
\multicolumn{1}{|l|}{1B Word dataset}          & Peters et al. 2017 \cite{peters2017semi}& \multicolumn{2}{l|}{96.37 $\pm$0.05}        \\ \hline
\multirow{8}{*}{None}                          & S{\o}gaard et al. 2016 \cite{sogaard2016deep} & \multicolumn{2}{l|}{95.28}                  \\ \cline{2-4} 
                                               & Yang et al. 2017 \cite{yang2017transfer} & \multicolumn{2}{l|}{94.66}                  \\ \cline{2-4} 
                                               & Hashimoto et al. 2017                   & \multicolumn{2}{l|}{95.02}                  \\ \cline{2-4} 
                                               & Rei 2017 \cite{rei2017semi} & \multicolumn{2}{l|}{93.88}                  \\ \cline{2-4} 
                                               & Peters et al. 2017 \cite{peters2017semi}  & \multicolumn{2}{l|}{95.00$\pm$0.08}         \\ \cline{2-4} 
                                              \cline{2-4}
                                               & \multirow{2}{*}{GTI-CHUNK-POS}          & \multicolumn{1}{c|}{mean} & 95.15$\pm$0.05 \\ \cline{3-4} 
                                               &                                         & \multicolumn{1}{c|}{max}  & 95.22           \\ \hline
\end{tabular}
}

\end{table}
As shown in Table \ref{tab:chunk_f1}, the performance of our GTI-CHUNK-POS model on the CoNLL-2000 Chunking dataset is comparatively lower than the state-of-the-art model \cite{peters2017semi}. 
The models by ~\cite{peters2017semi,hashimoto2017joint,sogaard2016deep} achieved high performance by using external resources/datasets. For example, the baseline model introduced by \cite{peters2017semi} achieved a score of $95.00$, approximately $1.37$ lower than when they expanded the training data with external corpus. 
The lower performance of the models trained on only the CoNLL-2000 chunking dataset can be attributed to the limited amount of available training data. Our GTI-CHUNK-POS model outperforms majority of the baselines trained without extra resources.
Similar to the GTI models trained on the CoNLL-2003 NER corpus, we observed that increasing the hidden state size (see Table \ref{tab:chunk_scores_state}) improves further the performance of the GTI-CHUNK-POS model but with a higher variance.
\subsection{Ablation}
\begin{table}[t!]
\centering
\caption{Ablation Results. The experiments are conducted on the CoNLL-2003 NER corpus with the LSTM state size is set to 200. The MTL models are trained with NER as the main-task and chunking and POS tagging as the auxiliary tasks. The STL models are trained to generate NER tags.}
\label{abl:result}
\scalebox{1}
{
\begin{tabular}{|l|c|c|c|}
\hline
\multicolumn{1}{|c|}{\multirow{2}{*}{Model}} & \multicolumn{3}{c|}{F$_1$}              \\ \cline{2-4} 
\multicolumn{1}{|c|}{}                       & Min   & Mean$\pm $ \textbf{std} & Max   \\ \hline
\multicolumn{4}{|c|}{Single Task Learning (STL) models}                                \\ \hline
BiLSTM-CRF(1)                               & 90.75 & 90.85$\pm$0.084         & 90.91 \\ \hline
BiLSTM-CRF(2)                               & 90.69 & 90.83$\pm$0.08          & 90.91 \\ \hline
\multicolumn{4}{|c|}{Multi-task Learning (MTL) models}                                 \\ \hline
Vanilla MTL                                  & 90.99 & 91.10$\pm$0.1           & 91.23 \\ \hline
Pipeline MTL                                 & 90.73 & 90.81$\pm$0.08          & 90.94 \\ \hline
TI-NER-CHUNK-POS                             & 90.93 & 91.13$\pm$0.11          & 91.26 \\ \hline
GTI-NER-CHUNK-POS                            & 91.27 & 91.38$\pm$0.07          & 91.49 \\ \hline
\end{tabular}
}
\end{table}
To investigate the impact of the \emph{Gated Interaction Layer (GIL)} in our proposed GTI approach for the generation of the NER tags, we performed 5 ablation studies.
The first 2 (BiLSTM-CRF(1) and BiLSTM-CRF(2)) are regular single-task learning baseline (Non-MTL) models for generating the NER tags. BiLSTM-CRF(1) is similar to the approach proposed by \cite{lample2016neural,ma2016end} for the task of NER. They trained a BiLSTM-CRF architecture using the words and character-level representation as inputs. BiLSTM-CRF(2) extends the words and character-level representations with linguistic features (chunking and the POS tags) as additional inputs to the model for the NER prediction. 

Three re-implementations of the GTI multi-task learning approach are also presented.
Vanilla MTL is the re-implementation without the \emph{Gated Interaction Layer (GIL)}. This model simultaneously learns the generation of the NER tasks along with the linguistic features (chunking and the POS tags). 
Pipeline MTL is a variant of  MTL model analogous to the BiLSTM-CRF(2) model and it is a  pipeline system which predicts the labels for auxiliary tasks (chunking and POS tag generation) first and then uses them as features for the main task (NER tagging). 
Without the GIL components (Trans\textsuperscript{k}, Composite\textsuperscript{k} and Gate\textsuperscript{k}), our proposed architecture is more similar Pipeline MTL. 
Finally, Task Interaction-NER-CHUNK-POS (TI-NER-CHUNK-POS) is also modeled after the  GTI-NER-CHUNK-POS model without the Gate\textsuperscript{k}  within the \emph{GIL}. 
TI-NER-CHUNK-POS's  \emph{Gated Interaction Layer} consists of only the Lemb\textsuperscript{k}, Trans\textsuperscript{k}, Composite\textsuperscript{k} and the Sum-block sub-layers. 
We compare the performance of the TI-NER-CHUNK-POS to GTI-NER-CHUNK-POS to verify that the performance gain over the Vanilla MTL and Pipeline MTL models is as a result of the gating mechanism controling the flow of information between the tasks.

As shown in Table \ref{abl:result}, we observed no significant change in performance when we augmented the input features with the linguistic features. Learning the linguistic features (chunking and the POS tags) simultaneously along with NER tagging under the Vanilla MTL approach yielded a marginal performance gain of about $0.25$ over the single models, BiLSTM-CRF(1) and BiLSTM-CRF(2).
The performance gain over the single models can be attributed to the inductive bias information provided by the auxiliary tasks \cite{Ruder2017AnOO}.
This enables the model to learn a shared representation beneficial to the main task.

The Pipeline MTL was expected to further improve the performance of the Vanilla MTL but surprisingly, the F$_1$ score dropped by about $0.29$ almost similar to the performance of the non-MTL models (BiLSTM-CRF(1) and BiLSTM-CRF(2)).   
 With the addition of the Trans\textsuperscript{k} and Composite\textsuperscript{k} components within the GIL, the Task Interaction model (TI-NER-CHUNK-POS)  obtained a higher performance compared to the Pipeline MTL model. But without the Gate\textsuperscript{k} component within the \emph{Gated Interaction Layer}, the TI-NER-CHUNK-POS model achieved no significant performance gain (F$_1$ score of $91.13\pm0.11$) over the Vanilla MTL model (F$_1$ score of $91.10\pm0.1$). 
 Adding the Gate\textsuperscript{k} component further improves the performance achieving a gain of about $0.57$, $0.28$ and $0.25$ higher than the Pipeline MTL, Vanilla MTL and TI-NER-CHUNK-POS  models respectively. The improvement in the F$_1$ score over the TI-NER-CHUNK-POS  model comes at no increase in  the number of hyperparameters of the model.
As mentioned above, the gating mechanism  enhances the interaction between the main task and its associated auxiliary tasks by controlling the information flow between  tasks under consideration.
Overall, the MTL models consistently outperform the single models (BiLSTM-CRF(1) and BiLSTM-CRF(2)) and the use of our gating mechanism via the \emph{Gated Interaction Layer} can  further improve the performance on the sequence generation task.

\section{Related Works}\label{sec:related_works}
Linguistic sequence labelling tasks such as NER, Chunking, SRL, POS etc, have been well studied over the years. 
But several of these approaches  such as SVM \cite{Kazama:2002:TSV:1118149.1118150}, CRF \cite{lafferty2001conditional} and Hidden Markov Models (HMMs) \cite{zhou2002named} depend on hand-crafted features. 
Models adopting hand-crafted features are difficult to adopt for a new domain or task.
For example, to adapt a NER model for POS tagging, new hand-crafted features have to be generated. This is a very expensive approach to learning.\\
\indent The automatic feature engineering via neural models has recently received a lot of attention. \cite{collobert2011natural} used CNN to extract input features vectors from the sequence of words. 
These features are then passed to CRF classifier. Contrast to this, other works by \cite{lample2016neural,he2017deep} replaced CNN with LSTM/BiLSTMs for extracting features from the input words. 
Also, \cite{ma2016end} combined  both LSTM and CNN. \cite{huang2015bidirectional} extended the input word-level feature vectors with hand-crafted features such as spelling features (e.g it whether starts with a capital letter, whether it has all lower case letters \etc) for a given word in the input sequence. 
Aside from the word-level features, recent works \cite{ma2016end,lample2016neural,chiu2016named} rely on the character-level features. 
All these approaches achieved remarkable performance over the traditional approaches to sequence labelling. 
To further improve the performance of neural models, other works  \cite{luo2017attention,sienvcnik2015adapting,marcheggiani2017encoding}  leverage additional information from linguistic features such as chunking and POS tags. However, the change in performance depends on the degree of semantic relatedness between the linguistic features and the target task.\\ 
\textbf{Multi-task Learning (MTL):} The downside to the application of deep learning is that it is data intensive requiring a large amount of labelled examples. To tackle the data scarcity problem, works by \cite{chiu2016named,collobert2011natural} augmented the dataset with language specific resources such as gazetters. Furthermore, recent approaches suggest guiding the learning process with extra knowledge via multi-task learning (MTL). Following the work by \cite{caruana1998multitask}, multi-task learning has been applied to many NLP problems as well as other neural network architectures.
The MTL approach to learning involves optimising multiple tasks simultaneously. 
This mostly involves sharing the model's weights between the tasks under consideration. Under the Sequence tagging, a number of MTL models have been proposed. \cite{peng2017multi} explored a shared representation learning strategy that supports domain adaptation for multiple tasks. \cite{bingel2017identifying} explored identifying the beneficial auxiliary tasks that can be modelled along with a given main/target task. Also,
\cite{collobert2008unified} proposed a unified network where the weights of the word embedding layer is shared between multiple sequence labelling tasks such as POS, SRL, Chunking and NER. \cite{sogaard2016deep} presented an approach to MTL where the different objectives are supervised at different levels. 
A similar approach was adopted by \cite{hashimoto2017joint} where they successively grew their network depth to tackle increasingly complex NLP tasks. 
In contrast to these works, the outputs of our auxiliary tasks are fed-back into the network via a \emph{Gated Interaction Layer (GIL)} which transforms them into features usable by the other tasks under consideration. This approach controls the flow of information between the multiple tasks.\\
\indent A distinction between the different MTL approaches also lies in the training strategy employed. 
A common approach involves training the MTL model on different task specific corpus by randomly switching between the different tasks and updating both the task-specific and shared parameters based on its corpus. 
\cite{alonso2017multitask,bingel2017identifying,hashimoto2017joint}~ employed this training strategy. 
A joint end-to-end model training strategy is mostly suitable for cases where the alternative tasks are treated as auxiliary objectives on the same dataset. 
Works by \cite{liu2017empower,peters2017semi,rei2017semi} trained a sequence labelling task such as NER along with unsupervised learning  tasks such as language modelling. 
The joint training approach eliminates the need for external corpus for the auxiliary tasks as the same dataset is used to learn all the multiple tasks.
We employed this strategy to train all the GTI models. 
\section{Conclusion}\label{sec:conc}
In this work, we proposed an MTL framework for sequence labelling, which exploits the relatedness between a given main/target task and its associated auxiliary tasks. 
The experimental results show that by jointly learning the linguistic features along with the target sequence labelling task, the GTI model achieves high performance on the baseline datasets for chunking and NER tagging tasks. 
The main and auxiliary tasks are trained on the same dataset eliminating the need for extra corpus for each task.
This makes the GTI framework ideal for low resource tasks such as NER and Chunking.

In the future, we aim to improve the performance of our GTI framework~ by exploring further~ the impact of other hyperparameters such as the dropout rates and number of BiLSTM layers. Finally, we intend to test the performance of our proposed approach on learning other sequence labelling tasks such as SRL and Error Detection. 
\section*{Acknowledgment}
This work is partially funded by the VCRS scholarship, by the EU Horizon 2020 Research and Innovation Programme under Grant 690238 for DESIREE Project, by the UK EPSRC under Grant EP/P031668/1,  by the BT Ireland Innovation Centre (BTIIC).
\bibliographystyle{IEEEtran}
\bibliography{main_script}
\end{document}